\documentstyle[ml99, epsfig, amsmath]{article}

\title{Learning Policies with External Memory}
\author{ {\bf Leonid Peshkin} \\  
Computer Science Dept. \\  
Brown University, Box 1910\\ 
Providence, RI, 02912-1910\\ 
ldp@cs.brown.edu \\
\And 
{\bf Nicolas Meuleau} \\ 
Computer Science Dept. \\ 
Brown University \\              
nm@cs.brown.edu\\
\And 
{\bf Leslie Pack Kaelbling}   \\ 
Computer Science Dept. \\ 
Brown University \\ 
lpk@cs.brown.edu\\ 
} 
 
\newcommand{\ql}{{\sc ql}}
\newcommand{\pomdp}{{\sc pomdp}}
\newcommand{\mdp}{{\sc mdp}}
\newcommand{\sarsa}{{\sc sarsa}}
\newcommand{\sarsal}{{\sc sarsa$(\lambda)$}}
\newcommand{\vaps}{{\sc vaps}}

\begin{document}

\maketitle

\begin{abstract}
In order for an agent to perform well in partially observable domains, it is
usually necessary for actions to depend on the history of
observations.  In this paper, we explore a {\it stigmergic} approach,
in which the agent's actions include the ability to set and clear bits
in an external memory, and the external memory is included as part of
the input to the agent.  In this case, we need to learn a reactive
policy in a highly non-Markovian domain.  We explore two algorithms:
{\sc sarsa}$(\lambda)$, which has had empirical success in partially
observable domains, and {\sc vaps}, a new algorithm due to Baird and
Moore, with convergence guarantees in partially observable domains.
We compare the performance of these two algorithms on benchmark problems.
\end{abstract}

\section{Introduction}

A reinforcement-learning agent must learn a mapping from a stream of
observations of the world to a stream of actions.  In completely
observable domains, it is sufficient to look only at the last
observation, so the agent can learn a ``memoryless'' mapping from
observations to actions \cite{Puterman94}. In general, however, the agent's actions
may have to depend on the history of previous observations.

\paragraph{Previous Work}

There have been many approaches to learning to behave in partially
observable domains.  They fall roughly into three classes: optimal
memoryless, finite memory, and model-based.

The first strategy is to search for the best possible memoryless
policy.  In many partially observable domains, memoryless policies can
actually perform fairly well.
Basic reinforcement-learning techniques, such as
Q-learning~\cite{Watkins92},  often perform poorly in partially
observable domains, due to a very strong Markov assumption.
Littman showed~\cite{LittmanICML94} that finding the optimal memoryless
policy is NP-Hard.  However, Loch
and Singh~\cite{Loch98} effectively demonstrated that techniques, such as 
\sarsal, that are more oriented toward optimizing total
reward, rather than Bellman residual, often perform very well.  In
addition, Jaakkola, Jordan, and Singh~\cite{Jaakkola94a} have developed
an algorithm for finding stochastic memoryless policies, which can
perform significantly better than deterministic ones \cite{Singh94a}.

One class of finite memory methods are the finite-horizon memory methods,
which can choose actions based on a finite window of previous observations.
For many problems this can be quite effective~\cite{McCallum95a,Ring94}. More
generally, we may use a finite-size memory, which can possibly be
infinite-horizon (the systems remembers only a finite number of events, but
these events can be arbitrarily far in the past). Wiering and
Schmidhuber~\cite{Wiering97} proposed such an approach, learning a policy
that is a finite sequence of memoryless policies.

Another class of approaches assumes complete knowledge of the
underlying process, modeled as a {\em partially observable Markov
  decision process} (\pomdp).  Given a model, it is possible to
attempt optimal solution~\cite{Kaelbling98}, or to search for
approximations in a variety of ways~\cite{Han98,Han98a,Hauskrecht98,Meu99}.
These methods can, in principle, be coupled with techniques, such as
variations of the Baum-Welch algorithm~\cite{Rabiner89}, for learning
the model to yield model-based reinforcement-learning systems.

\paragraph{Stigmergy}

In this paper, we pursue an approach based on {\em stigmergy}.  The
term is defined in the Oxford English Dictionary~\cite{OED} as ``The
process by which the results of an insect's activity act as a stimulus
to further activity,'' and is used in the mobile robotics
literature~\cite{Beckers94} to describe activity in which an agent's
changes to the world affect its future behavior, usually in a useful
way.

One form of stigmergy is the use of external memory devices.  We are
all familiar with practices such as making grocery lists, tying a
string around a finger, or putting a book by the door at home so you
will remember to take it to work.  In each case, an agent needs to
remember something about the past and does so by modifying its 
external perceptions in such a way that a memoryless policy will
perform well.

\begin{figure}[tbp]
        \centerline{\epsfig{file=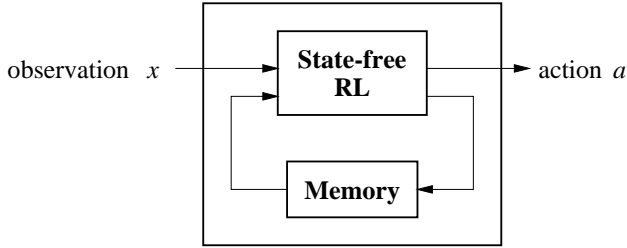,width=3.25in}}
        \caption{The architecture of a stigmergic policy.}
        \label{arch}
\end{figure}

We can apply this approach to the general problem of learning to
behave in partially observable environments.  Figure~\ref{arch} shows
the architectural idea.  We think of the agent
as having two components: one is a set of memory bits;  the other is
a reinforcement-learning agent. The reinforcement-learning agent has
as input the observation that comes from the environment, augmented by
the memory bits. Its output consists of the original actions in the
environment, augmented by actions that change the state of the
memory. If there are sufficient memory bits, then the optimal
memoryless policy for the internal agent will cause the entire agent
to behave optimally in its partially observable domain.

\begin{figure}[tbp]
        \centerline{\epsfig{file=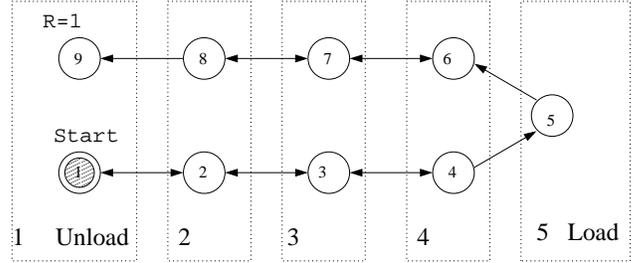,width=3.25in}}
        \caption{The state-transition diagram of the load-unload problem;
           aliased states are grouped by dashed boxes.}
        \label{loadunload}
\end{figure}

Consider, for instance, the load-unload problem represented in
Figure~\ref{loadunload}. In this problem, the agent is a cart that must drive
from an {\em Unload} location to a {\em load} location, and then back to {\em
unload}. This problem is a simple MDP with a one-bit hidden variable that
makes it non-Markov (the agent cannot see whether it is loaded or not). It
can be solved using a one-bit external memory: we set the bit when we make
the {\em Unload} observation, and we go right as long as it is set to this
value and we do not make the {\em Load} observation. When we do make the {\em
Load} observation, we clear the bit and we go left as long as it stays at
value 0, until we reach state 9, getting a reward.

There are two alternatives for designing an architecture with external
memory:
\begin{itemize}
\item Either we {\em augment} the action space with actions that change the content of one of the memory bits (adds $L$ new actions if there are L memory
  bits); changing the state of the memory may require multiple steps. 
\item Or we {\em compose} the action space with the set of all possible
  values for the memory (the size of the action space is then multiplied by
  $2^L$, if there are $L$ bits of memory). In this case, changing the
  external memory is an instantaneous action that can be done at each
  time step in parallel with a primitive action, and hence we can
  reproduce the optimal policy of the load-unload problem, without taking
  additional steps. 
\end{itemize}
Complexity considerations usually lead us to take the first option. It
introduces a bias, since we have to lose at least one time-step each time we
want to change the content of the memory. However, it can be fixed in most
algorithms by not discounting memory-setting actions.

The external-memory architecture has been pursued in the context of classifier systems \cite{Booker89} and  in the context of
reinforcement learning by Littman~\cite{LittmanICML94} and by
Mart\'\i n~\cite{Martin98}.  Littman's work was model-based; it assumed
that the model was completely known and did a branch-and-bound search
in policy space. Mart\'\i n worked in the model-free
reinforcement-learning domain; his algorithms were very successful at
finding good policies for very complex domains, including some
simulated visual search and block-stacking tasks.  However, he made a
number of strong assumptions and restrictions: task domains are
strictly goal-oriented; it is assumed that there is a deterministic
policy that achieves the goal within some specified number of steps
from every initial state; and there is no desire for optimality in
path length.

\paragraph{This work}

We were inspired by the success of Mart\'\i n's algorithm on a set of
difficult problems, but concerned about its restrictions and a number
of details of the algorithm that seemed relatively {\em ad hoc}.  At
the same time, Baird and Moore's work on \vaps~\cite{Baird99}, a
general method for gradient descent in reinforcement learning,
appealed to us on theoretical grounds.  This paper is the result of
attempting to apply \vaps~algorithms to stigmergic policies, and
understanding how it relates to Mart\'\i n's algorithm.  In this process,
we have derived a much simpler version of \vaps~for the case of highly
non-Markovian domains: we calculate the same gradient as \vaps, but
with much less computational effort.

In the next section, we present the relevant learning algorithms.
Then we describe a set of experimental domains and discuss the
relative performance of the algorithms.

\section{Algorithms}

We begin by describing the most familiar of the algorithms, \sarsal.
We then describe the \vaps~algorithm in some detail, followed by our
simplified version.

\subsection{\sarsal}

\sarsa~is an on-policy temporal-difference control learning
algorithm~\cite{Sutton98a}. Given an experience in the world, characterized
by starting state $x$, action $a$, reward $r$, resulting state $x'$ and next
action $a'$, the update rule for \sarsa(0) is 
\begin{equation}
Q(x,a) \leftarrow Q(x,a) + \alpha [r + \gamma Q(x',a') - Q(x,a)]\;.
\label{sarsa}
\end{equation}
It differs from the classical Q-learning algorithm \cite{Watkins89} in that, rather than using the maximum $Q$-value from the resulting state as an estimate of that state's value,
it uses the $Q$-value of the resulting state and the action that was
actually chosen in that state. Thus, the values learned are sensitive
to the policy being executed.

In truly Markov domains, Q-learning is usually the algorithm of choice;
policy-sensitivity is often seen as a liability, because it makes issues of
exploration more complicated. However, in non-Markov domains,
policy-sensitivity is actually an asset. Because observations do not uniquely
correspond to underlying states, the value of a policy depends on the
distribution of underlying states given a particular observation. But this
distribution generally depends on the policy. So, the value of a state, given
a policy, can only be evaluated {\em while executing that policy}. In fact, Q-learning can be shown to fail to converge on very simple non-Markov
domains~\cite{Singh94}. Note that, when \sarsa~is used in a non-Markovian environment, the symbols $x$ and $x'$ in equation (\ref{sarsa}) represent observations, which usually can correspond to several states.

The \sarsa~algorithm can be augmented with an eligibility trace, to
yield the \sarsal~algorithm (a detailed exposition is given by Sutton
and Barto~\cite{Sutton98a}.)  With the parameter $\lambda$ set to 0,
\sarsal~is just \sarsa. With $\lambda$ set to 1, it is a pure Monte
Carlo method, in which, at the end of every trial, each state-action
pair is adjusted toward the cumulative reward received on this trial
after the state-action pair occurred. Pure Monte-Carlo algorithms
make no attempt at satisfying Bellman equations relating the values of
subsequent states;  in partially observable domains, it is often
impossible to satisfy the Bellman equation, making Monte-Carlo a
reasonable choice.  \sarsal~describes a useful class of algorithms,
then, with appropriate choice of $\lambda$ depending on the problem.
Thus, \sarsal~with a large value of $\lambda$ seems like the most appropriate of the conventional
reinforcement-learning algorithms for solving partially-observable problems.

\subsection{\vaps}

Baird and Moore have derived, from first principles, a class of stochastic
gradient-descent algorithms for reinforcement learning.
At the most abstract level, we seek to minimize some measure of the
expected cost of our policy; we can describe this high-level
criterion as
\[B = \sum_{T = 0}^\infty \sum_{\tilde{s} \in \tilde{S}_T} \Pr(\tilde{s}) \varepsilon(\tilde{s})\;\;,\]  
where $\tilde{S}_T$ is the set of all possible experience sequences that
terminate at time $T$.  That is, 
$$\tilde{s} =\langle x_0,u_0,r_0,...,x_t,u_t,r_t, ...,x_T,u_T,r_T
\rangle\;,$$ where $x_t$, $u_t$, and $r_t$ are the observation,
action, and reward at step $t$ of the sequence, and $x_T$ is an observation associated with a terminal state.  The loss incurred by a
sequence $\tilde{s}$ is $\varepsilon(\tilde{s})$. We restrict our attention to time-separable
loss functions, which can be written as
$$\varepsilon(\tilde{s}) = \sum_{t = 0}^T e({\rm trunc}(\tilde{s}, t))\;,\hspace{1cm}{\rm for\ all\ }\tilde{s}
\in \tilde{S}_T,$$ where $e(s)$ is an instantaneous error function
associated with each (finite) sequence prefix $s =\langle
x_0,u_0,r_0,...,x_t,u_t,r_t\rangle$ ($x_t$ being any observation, not
necessarily a terminal one), and ${\rm trunc}(\tilde{s}, t)$ representing the
sequence $\tilde{s}$ truncated after time $t$. For instance, an error measure closely
related to Q-learning is the squared Bellman residual:
\begin{equation*}
\begin{split}
e_{\ql}(s) = & \frac{1}{2}\sum_x \Pr(x_{t} = x \mid
x_{t-1},u_{t-1}) \\
& [r_{t-1} + \max_u \gamma Q(x,u) -
Q(x_{t-1},u_{t-1})]^2\;.\\
\end{split}
\end{equation*}
The \sarsa~version of the algorithm uses the following error measure:
\begin{equation*}
\begin{split}
& e_{\sarsa}(s) =\\
& \quad \frac{1}{2}\sum_x \Pr(x_{t} = x \mid
x_{t-1},u_{t-1}) \sum_{u}\Pr(u_t = u \mid x_t)\\
&\quad [r_{t-1} + \gamma
Q(x,u) - Q(x_{t-1},u_{t-1})]^2\;.
\end{split}
\end{equation*}
Note that we average over all possible actions $u_t$ according to their probability of being chosen by the policy
instead of picking the one that maximizes Q-values as in $e_{\ql}$. Baird and Moore also
consider a kind of policy search, which is analogous to {\sc
  reinforce}~\cite{Williams87}:
\[ e_{\rm policy}(s) = b - \gamma^t r_t\;,\]
 where $b$ is any constant. This immediate error is
summed over all time $t$, leading to a summation of all discounted
immediate rewards $\gamma^t r_t$.  In order to obtain the good
properties of both criteria, they construct a final criterion that is
a linear combination of the previous two:
\[e = (1-\beta) e_{\sarsa} + \beta e_{\rm policy}\;.\]  This
criterion combines Value And Policy Search and is, hence, called
\vaps.  We will refer to it as \vaps$(\beta)$, for different values of
$\beta$.
       
Baird and Moore show that the gradient of the global error with
respect to weight $k$ can be written as:
\begin{equation*}
\begin{split}
& \frac{\partial}{\partial w_k} B = \sum_{t=0}^{\infty}\sum_{s \in S_t}
\Pr(s)\\
& \quad \left[\frac{\partial}{\partial w_k}e(s) +
e(s)\sum_{j=1}^t \frac{\partial}{\partial w_k}\ln\Pr(u_{j-1} |
x_{j-1})\right]\;,
\end{split}
\end{equation*}
where $S_t$ is the set of all experience prefixes of length $t$. Technically,
it is necessary that $\Pr(u^t = u \mid x_t = x) > 0$ for all $(x, u)$
(otherwise, some zero probability trajectories may have a non-zero
contribution to the gradient of $B$ \cite{Kaelbling99}). In this work we use
the Boltzmann law for picking actions, which guarantees this property (see
section 2.3).
 
It is possible to perform stochastic gradient descent of the ``error'' $B$, by repeating several trials of interaction with the
process. Each experimental trial of length $T$ provides one sample of $s \in
S_t$ 
for each $t \leq T$. Of course, these samples are not independent, but it does
not matter since we are summing them and not multiplying them. We are
thus using stochastic approximation to estimate the expectation over
$s \in S_t$ in the above equation. During each trial,
the weights are kept constant and the approximate gradients of the
error at each time $t$,
$$\frac{\partial}{\partial w_k}e(s) + e(s)\sum_{j=1}^t
\frac{\partial}{\partial w_k}\ln\Pr(u_{j-1} | x_{j-1})\;,$$ are
accumulated. Weights are updated at the end of each trial, using the sum
of these immediate gradients. An incremental implementation of the
algorithm can be obtained by using, at every step $t$, the following
update rules:
\begin{eqnarray*}
\Delta T_{k,t} & = & \frac{\partial}{\partial w_k}\ln\Pr(u_{t-1} | x_{t-1})\;, \\
\Delta w_k & = & - \alpha \left[ \frac{\partial}{\partial w_k}e(\tilde{s}_t)
+ e(\tilde{s}_t) T_{k,t} \right]\;,
\end{eqnarray*}
where $s_t$ represents the experience prefix \\ $\langle x_0,u_0,r_0, \ldots,
x_t,u_t,r_t\rangle$, i.e., the history at time $t$. Note that the
``exploration trace'' $T_{k,t}$ is independent of the immediate error $e$
used. It only depends on the way the output $\Pr(u_t = u|x_t)$ varies with
the weights $w_k$, i.e., on the representation chosen for the policy.

The gradient of the immediate error $e$ with respect to the weight
$w_k$ is easy to calculate. For instance, in the case of the 
\sarsa~ variant of the algorithm we have:
\begin{equation*}
\begin{split}
&  \frac{\partial}{\partial w_k} e_{\sarsa}(s) =  \\
& \quad \sum_x
  \Pr(x_t = x \mid x_{t-1},u_{t-1}) \sum_{u}\Pr(u_t = u \mid x_t) \\ 
& \quad  \left[r_{t-1} + \gamma Q(x,u) - Q(x_{t-1},u_{t-1})\right] \\
& \quad \left[\gamma
  \frac{\partial}{\partial w_k}Q(x,u) - \frac{\partial}{\partial
    w_k}Q(x_{t-1},u_{t-1})\right]\;.
\end{split}
\end{equation*}
Once more, we descend this gradient by stochastic approximation: the
averaging over $x_t$ and $u_t$ is replaced by a sampling of these quantities.
However, since these variables appear twice in the equation and they are not
just added, we have to sample both $x_t$ and $u_t$ independently two times in
order to avoid any bias in the estimation of the gradient. 
It is not realistic to satisfy this requirement in a truly on-line situation,
since the only way to get a new observation is by actually {\it performing} the
action. Note that for the case $\beta =1$ we do not need the second sample,
so the \vaps(1) algorithm is effective in the on-line case. 

In the case of policy search we have:
$\frac{\partial}{\partial w_k} e_{\rm policy}(s)=0$, \\
for all $w_k$. This may seem strange;  but for policy search, the
important thing is the state occupations, which enter into the weight
updates through the trace.


\subsection{\vaps(1)}

In this section, we explore a special case of \vaps, in which the $Q$-values
are stored in a look-up table. That is, there is one weight $w_k = Q(x, u)$
for each state-action pair. Note that it is not necessary to use the
\vaps~sequence-based gradient in a look-up table implementation of \ql~or
\sarsa, as long as it is confined to a Markovian environment. However, it
makes sense to use it in the context of \pomdp s.
Under this hypothesis, the exploration trace $T_{k, t}$
associated with each parameter $Q(x, u)$ will be written $T_{x, u, t}$.

We will also focus on a very popular rule for randomly selecting
actions as a function of their $Q$-value, namely the Boltzmann law:
$$Pr(u_t = u| x_t = x)  = \frac{e^{Q(x,u)/c}}{\sum_{u'}
  e^{Q(x,u')/c}}\;,$$ 
where $c$ is a temperature parameter.\footnote{Note that Baird and Moore
  use an unusual version of the Boltzmann law, with $1+e^x$ in place
  of $e^x$ in both the numerator and the denominator.  We have found
  that it complicates the mathematics and worsens the performance, so
  we will use the standard Boltzmann law throughout.}
 Under this rule we get:
\begin{equation*}
\begin{split}
&\frac{\partial \ln\Pr(u_t = u| x_t = x)}{\partial Q(x', u')} = \\
& \left\{
\begin{array}{ll}
 0 & {\rm if\ }x' \neq x,\\
-\Pr(u_t = u'| x_t = x)/c & {\rm if\ }x' = x {\rm\ and\ }u' \neq u,\\
\left[1 -\Pr(u_t = u| x_t = x)\right]/c &{\rm if\ }x' = x {\rm\ and\ }u' = u.\\
\end{array}\right.\\
\end{split}
\end{equation*}

In this case, and if we add the hypothesis that the problem is an
achievement task, i.e., the reward is always 0 except when we reach an
absorbing goal state, the
exploration trace $T_{x, u, t}$ takes a very simple form:
\begin{equation}\begin{array}{lll}
T_{x,u,t} & = & {1 \over c} \left[N_{x,u}^t - N_x^t \Pr\left(u_t = u| x_t =
x\right)\right] \\
& = & {1 \over c}\left[N_{x,u}^t - E[N_{x,u}^t]\right]\;,
\label{eqn:ldp}
\end{array}\end{equation}
where $N_{x,u}^t$ is the number of times that action $u$ has been
executed in state $x$ at time $t$, $N_x^t$ is the number of times that
state $x$ has been visited at time $t$, and $E[N_{x,u}^t]$ represents
the expected number of times we should have performed action $u$ in
state $x$, knowing our exploration policy and our previous history. 

As a result of equation (\ref{eqn:ldp}), \vaps~using $e_{policy}$ as
immediate error, look-up tables and Boltzmann exploration reduces to a
very simple algorithm. At each time-step where the current trial does
not complete, we just increment the counter $N^t_{x, u}$ of the
current state-action pair. When the trial completes, this trace is
used to update all the $Q$-values, as described above.

It is interesting to try to understand the properties and implications of
this simple rule. First, a direct consequence is that when something
surprising happens, the algorithm adjusts the unlikely actions more than the
likely ones. In other words, this simple procedure is very intuitive, since
it assigns credit to state-action pairs proportional to the deviation from
the expected behaviour. Note that \sarsal~is not capable of such a
discrimination. This difference in behaviour is illustrated in the simulation
results.

A second interesting property is that the $Q$-value updates tend to $0$ as
the length of the trial tends to infinity. This also makes sense, since the
longer the trial, the less the final information received (the final reward)
is relevant in evaluating each particular action. Alternatively, we could say
that when too many actions have been performed, there is no reason to
attribute the final result more to one of them than to others. Finally,
unlike with Baird and Moore's version of the Boltzmann law, the sum of the
updates to the $Q$-values on every step is 0. This makes it more likely that
the weights will stay bounded.

\section{Experiments}

\begin{figure}[tbp]
        \centerline{\epsfig{file=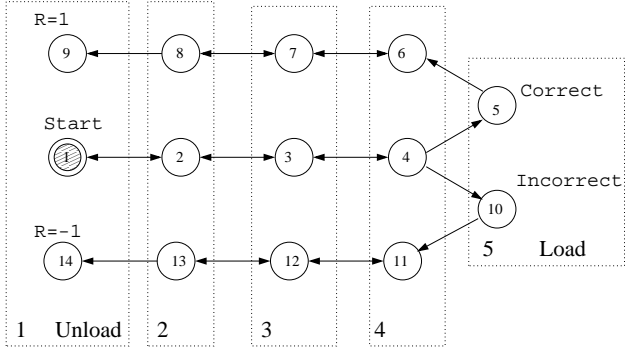,width=3.25in}}
        \caption{The state-transition diagram of the load-unload problem with two loading locations;
           aliased states are grouped by dashed boxes.}
        \label{LU 2loc}
\end{figure}

\paragraph{Domains}
We have experimented with \sarsa~and \vaps~on five simple problems.
Two are illustrative problems previously used in the
reinforcement-learning literature;  two others are instances of load-unload with different parameters; and the fifth is a variant of load-unload designed by us in an attempt to demonstrate a situation in which \vaps~might outperform
\sarsa.  The five problems are :
\begin{itemize}
\item Baird and Moore's problem~\cite{Baird99}, designed to illustrate
  the behavior of \vaps, 
\item McCallum's 11-state maze~\cite{McCallum95a}, which has only 6
  observations.
\item The load-unload problem, as described above, in which there are
  three locations (the loading location, the unloading location, and
  one intermediate one), 
\item A five-location load-unload problem (fig.~\ref{loadunload}), and 
\item A variant of the load-unload problem where a second loading location has been added, and the agent is punished instead of rewarded if it gets loaded at the wrong location. The state space is shown in
figure~\ref{LU 2loc}; states contained in a box are observationally
indistinguishable to the agent. The idea here is that there is a single action that, if chosen, ruins
the agent's long-term prospects.   If this action is chosen due to
exploration, then \sarsal~ will punish all of the action choices along
the chain but \vaps~ will punish only that action.
\end{itemize}
All these domains have a single starting state, except McCallum's
problem, where the starting state is chosen uniformly at random.

\paragraph{Algorithmic Details}

For each problem, we ran two algorithms: \vaps(1) and \sarsa(1). The optimal
policy for Baird's problem is memoryless, so the algorithms were applied
directly in that case. For the other problems, we augmented the input space
with an additional memory bit, and added two actions: one for setting the bit
and one for clearing it.

The $Q$-functions were represented in a table, with one weight for
each observation-action pair. The learning rate is determined by a parameter, $\alpha_0$;  the
actual learning rate has an added factor that decays to 0 over time:
$\alpha = \alpha_0 + {1 \over 10 N}$,
where $N$ is trial number. The temperature was also decayed in an {\em ad hoc} way, from $c_{\rm max}$
down to $c_{\rm min}$ with an increment of 
$$\delta c =
\left(\frac{c_{min}}{c_{max}}\right)^{1/(N-1)}$$
on each trial. In order to
guarantee convergence of \sarsa~in \mdp s, it is necessary to decay the
temperature in a way that is dependent on the $Q$-values
themselves~\cite{Singh99}; in the \pomdp~setting it is much less clear what
the correct decay strategy is. In any case, we have found that the empirical
performance of the algorithm is not particularly sensitive to the
temperature. The parameter $b$ in the immidiate error $e_{\rm policy}$ of
\vaps~ was always set to $0$.

\paragraph{Experimental Protocol}
Each learning algorithm was executed for $K$ runs;  each run consisted
of $N$ trials, which began at the start state and executed until a
terminal state was reached or $M$ steps were taken.  If the run
was terminated at $M$ steps, it was given a terminal reward of -1;
$M$ was chosen, in each case, to be 4 times the length of the optimal
solution. 
At the beginning of each run, the weights were randomly reinitialized
to small values.

\paragraph{Results}

\begin{figure}[tbp]
        \centerline{\epsfig{file=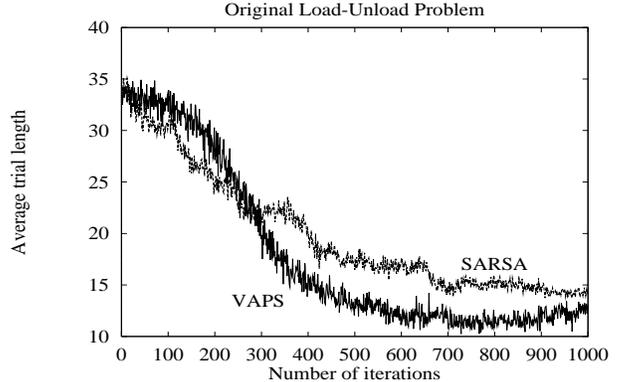,width=3.25in, height = 2in}}
        \caption{Learning curves for \vaps~  and \sarsa~  on the
          load-unload problem (one loading location).}
        \label{lugraph}
\end{figure}

It was easy to make both algorithms work well on the first three
problems:  Baird's, McCallum's and small load-unload.  The 
algorithms typically converged in fewer than 100 runs to an optimal
policy.  One thing to note here is that our version of \vaps, using
the true Boltzmann exploration distribution rather than the one
described by Baird and Moore, seems to perform significantly better
than the original, according to results in their paper.

Things were somewhat more complex with the last two problems (5 location load-unload with one or two loading locations).  We experimented
with parameters over a broad range and determined the following:
\begin{itemize}
\item \vaps~requires a value of $\beta$ equal or very nearly equal to
  1;  these problems are highly non-Markovian, so the Bellman error is
  not at all useful as a criterion.  
\item For similar reasons, $\lambda = 1$ is best for \sarsal.
\item Exploration was simplified by setting $\epsilon$ to 0;
  empirically, $c_{\it max}=1.0$
 and $c_{\it min}=0.2$ worked well for
  \vaps~in both problems, and $c_{\it max}=0.2$,  $c_{\it min}=0.1$
  worked well for \sarsal.
\item A base learning rate of $\alpha_0 = 0.5$ worked well for both
  algorithms in both domains.
\end{itemize}

\begin{figure}[tbp]
        \centerline{\epsfig{file=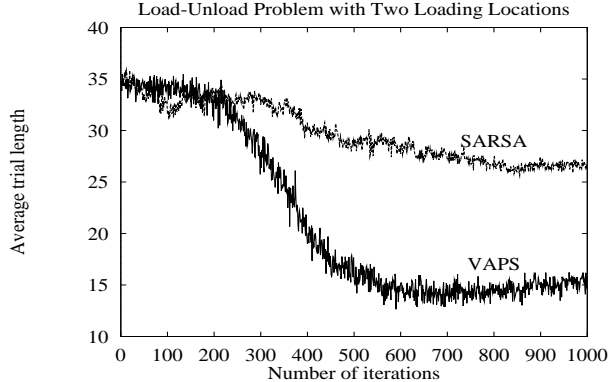,width=3.25in, height = 2in}}
        \caption{Learning curves for \vaps~ and \sarsa~ on the
          load-unload problem with two loading locations.}
        \label{shopgraph}
\end{figure}

Figures~\ref{lugraph} and \ref{shopgraph} show learning curves for
both algorithms, averaged over 50 runs, on the load-unload problem with one or two loading locations.  Each run consisted of 1,000 trials.  The vertical
axis shows the number of steps required to reach the goal, with the
terminated trials considered to have taken $M$ steps.

On the original load-unload problem, the algorithms perform essentially
equivalently. Most runs of the algorithm converge to the optimal trial length
of 9 and stay there; occasionally, however it reaches 9 and then diverges.
This can probably be avoided by decreasing the learning rate more steeply.
When we add the second loading location, however, there is a significant
difference. \vaps(1) consistently converges to a near-optimal policy, but
\sarsa(1) does not. The idea is that sometimes, even when the policy is
pretty good, the agent is going to pick up the wrong load due to exploration
and get punished for it. \sarsa~ will punish all the state-action pairs
equally; \vaps(1) will punish the bad state-action pair more due to the
different principle of credit assignment.

\section{Conclusions}

As Mart\'\i n and Littman showed, small \pomdp s can be solved effectively
using stigmergic policies. Learning reactive policies in highly non-Markovian
domains is not yet well-understood. We have seen that the \vaps~algorithm,
somewhat modified, can solve a collection of small \pomdp s, and that
although \sarsal~performs well on some \pomdp s, it is possible to construct
cases on which it fails. In a generalization of this work, we applied the
\vaps~algorithm to the problem of learning general finite-state controllers
(which encompass external-memory policies) for \pomdp s~\cite{Meu99a}.

\subsubsection*{Acknowledgments} 
 
This work was supported in part by DARPA/Rome Labs Planning Initiative
grant F30602-95-1-0020.
 
\bibliographystyle{plain}
\bibliography{/home/ai/pesha/BiB/lpk,/home/ai/pesha/BiB/icml99,/home/ai/pesha/BiB/thesis}
\end{document}